\documentclass[12pt,a4paper]{article} %

\usepackage[cp1250]{inputenc}
\usepackage{polski}
\usepackage{amssymb,amsmath} 
\usepackage{hyperref}
\usepackage{graphicx}
\usepackage{titlesec}
\usepackage{subcaption}
\usepackage{color}
\usepackage{algorithm}
\usepackage[numbers]{natbib}
\usepackage{float} 
\usepackage{footnote}
\usepackage{listings}
\usepackage{lipsum, enumitem}
\usepackage{xcolor}
\usepackage{multirow}
\usepackage{rotating}
\usepackage{subcaption}
\usepackage{algorithmicx}
\usepackage{algpseudocode}

\algnewcommand{\algorithmicand}{\textbf{ and }}
\algnewcommand{\algorithmicor}{\textbf{ or }}
\algnewcommand{\OR}{\algorithmicor}
\algnewcommand{\AND}{\algorithmicand}
\algnewcommand{\var}{\texttt}

\title{The recursive scheme of clustering}

\author{\small {Alicja Miniak-G{\'o}recka 0000-0002-1860-8853} \and
\small {Krzysztof Podlaski 0000-0002-2883-0773} \and
\small {Tomasz Gwizda{\l}{\l}a 0000-0002-3981-6037} \and
\footnotesize {Faculty of Physics and Applied Informatics} \and
\footnotesize {University of {\L}\'od\'z} \and 
\footnotesize {Pomorska 149/153, 90-236 {\L}\'od\'z, Poland} \and
\footnotesize {alicja.miniak, krzysztof.podlaski, tomasz.gwizdalla@uni.lodz.pl}}

\date{}

\begin{document}
\maketitle

\thispagestyle{empty}

\section{Abstract}
The problem of data clustering is one of the most important in data analysis. It can be problematic when dealing with experimental data characterized by measurement uncertainties and errors. Our paper proposes a recursive scheme for clustering data obtained in geographical (climatological) experiments. The discussion of results obtained by k-means and SOM methods with the developed recursive procedure is presented. We show that the clustering using the new approach gives more acceptable results when compared to experts' assessments.

\section{Introduction}

The clustering of experimental data is usually performed at the beginning stage of data analysis. Searching for the possibility of grouping the observations according to some similarities is the basic way to organize the data. That is why the clustering processes are widely studied, and the number of different approaches grows yearly. One can find the comparison of selected methods, e.g., in \cite{rodriguez,everitt}.

In our paper, we propose the recursive scheme of clustering based on the analysis of histograms smoothed by the Savitzky-Golay algorithm. The method extends the popular techniques, like k-means or SOM, by introducing the limitations on the number of clusters determined during one step of the procedure and further divisions of thus obtained clusters. The proposed methods have some advantages when compared with the pure formulation of seminal methods. There are mainly two main points that can be mentioned here. The first one is the decreased dependence of the result on the initial choice of centers. Another essential property is the concurrent performance of two processes: the division and the determination of the number of clusters. 
Our proposition belongs to the class of multi-phase approaches which, in different contexts and formulations, appear recently frequently \cite{Farshidvard_23,Hadi_22,Sui_22,Colak_22}

The results will be shown on three datasets. The first two are the raw results of climatological measurements led by the group from the University of Lodz in the northeastern wetlands of Poland \cite{fortuniak}. There are several reasons justifying such a choice. The data are hard to analyze due to significant measurement uncertainties; the climatological data are today an interesting and widely studied object of study (see, e.g., \cite{Carro-Calvo_21,Prakaisak_22}). The third, very important reason is that in discussion with colleagues involved in the study, we can retrieve particular behavior of studied parameters - we will call it contact with an expert. The third dataset is the Banknote Authentication sample from Machine Learning Repository \cite{mlr}. The dataset is also the object of interest of different studies \cite{Jadhav_19,Alguliyev_20}. However, multidimensional clustering is mainly under consideration.

\section{Method and results}\label{mar}

As a result of cooperation with the climate researchers group, we can access raw, real data from experiments conducted at the Biebrza National Park's wetlands in northeastern Poland.
Observational data on greenhouse gas exchange between ecosystems and the atmosphere is crucial in understanding the global climate mechanisms. Still, such data are highly scarce for Central European wetlands \cite{fortuniak}.
Therefore, the paper presents the study considering the selected data from continuous (2013-2017) measurements of fifteen climatological parameters (e.g., methane, carbon dioxide fluxes, ground temperature, water level, and wind direction).
As the authors suggest, the data is burdened with significant measurement uncertainty and errors \cite{fortuniak}.

\subsection{The problem}

\begin{figure}[!h!]
\centering
\includegraphics[width=1.0\columnwidth]{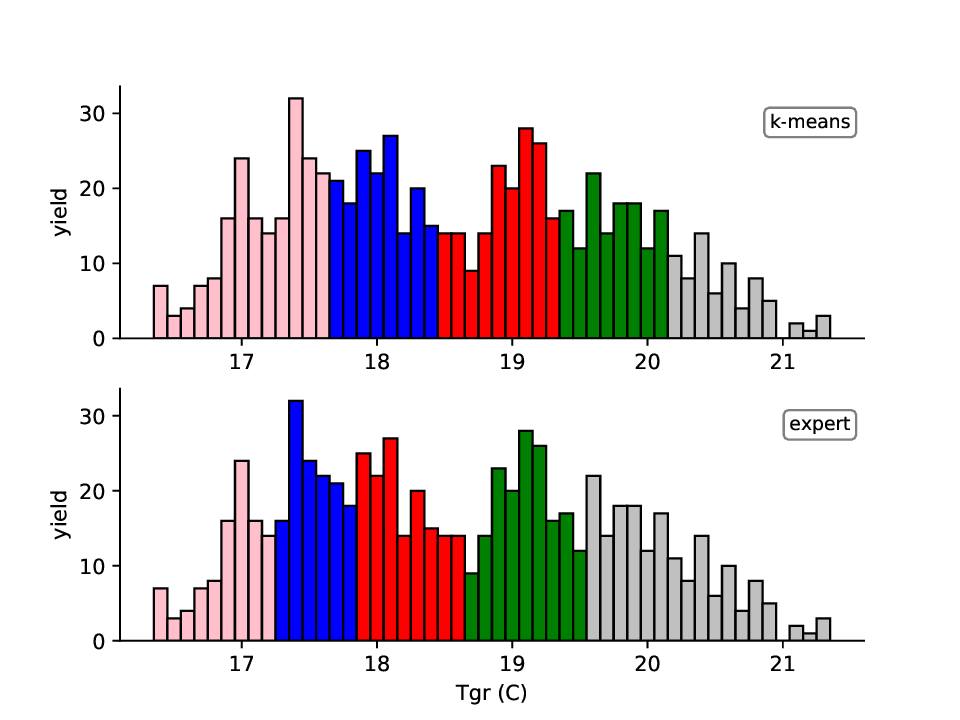}
\caption{Comparison of two different divisions of ground temperature data into five clusters using the classical k-means method and the expert's decision.}
\label{fig:two_histograms}
\end{figure}

There are many existing methods of clustering. For our research, we choose two well-known out-of-the-box approaches, k-means and SOM. The data to be clustered is connected with climatological measurements. First, let us look more carefully at ground temperature measurements. 
Experts suggested to divide the data into five clusters. We use k-means clustering to the same data and compare it with experts' choice (Fig. \ref{fig:two_histograms}).  

Experts' divisions and k-means clusters are different. We can easily see that experts divide all range of values into two parts with a border near point 18.7 and next to divide both parts independently. In contrast, the k-means method does not discover a similar structure and divides all results into five clusters with similar subranges. This example proves that the existing clustering methods do not always give the expected results. Thus, we decided to modify a clustering approach. We incorporate existing k-means and SOM algorithms with a recursive scheme as our experts' approach.

The problem of division into clusters has one fundamental issue that seriously impacts the clustering results. The user has to give an appropriate number of clusters, and most techniques that help to determine the number of clusters have a heuristic character.
They are usually based on the typical distance analysis or the comparison to some assumed distribution, even when some mathematical apparatus stays behind them, e.g., in the Silhouette method \cite{Rousseeuw_87} or Gap statistics \cite{Tibishirani_01}. The popular elbow technique was also often criticized for its ambiguity \cite{Ketchen_96}. 

\begin{figure}[!ht]
\centering
\begin{subfigure}[b]{0.8\columnwidth}
    \includegraphics[width=\columnwidth]{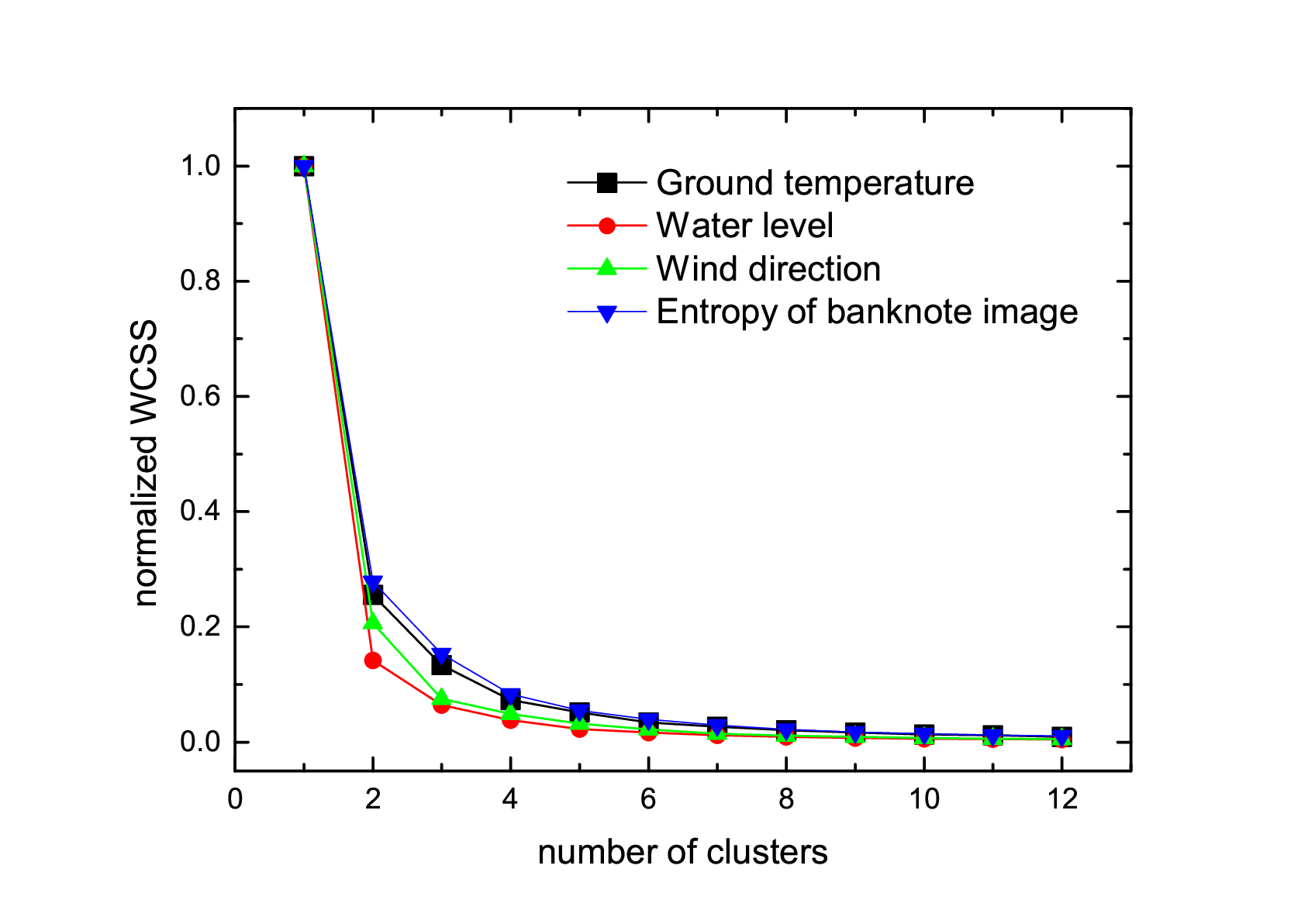}
    \caption{Elbow method.}
    \label{elbow}
\end{subfigure}
\begin{subfigure}[b]{0.8\columnwidth}
    \includegraphics[width=\columnwidth]{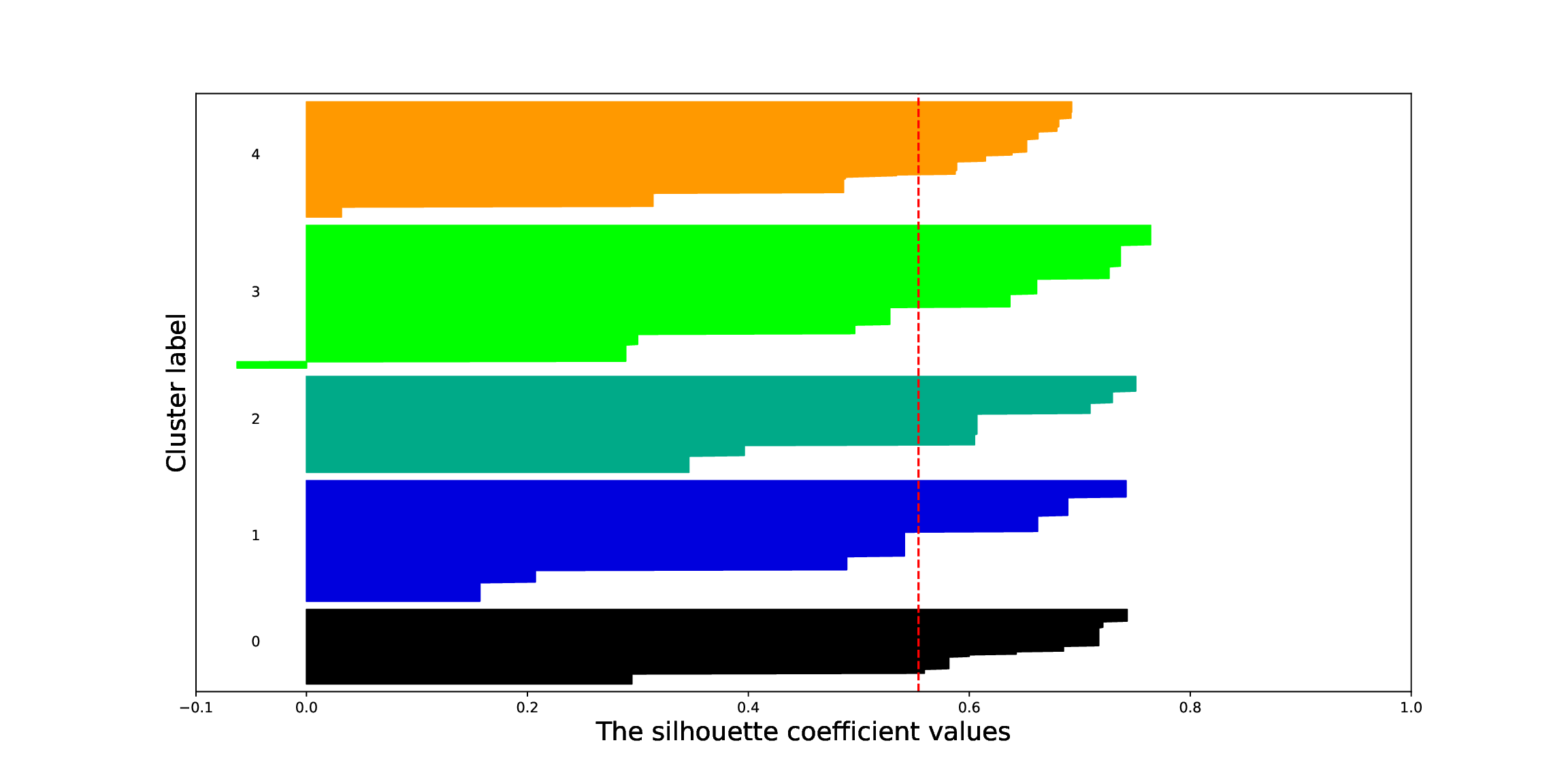}
    \caption{Silhouette method.}
    \label{sil_tgrunt}
\end{subfigure}
\caption{The silhouette and elbow methods applied to ground temperature data.}
\end{figure}

Let us look at the ground temperature data using elbow and silhouette methods. The elbow methods suggest dividing data into clusters of two, three, or four (Fig.\ref{elbow}). The silhouette method also does not clearly indicate a specific number of clusters. In Figure \ref{sil_tgrunt}, we present only silhouette visualization for five clusters, but silhouettes for two, three, and four clusters look very similar. As a result, both methods give an ambiguous answer on the best number of clusters for the data.

\subsection{Savitzky-Golay recursive clustering scheme}

The recursive approach requires data analysis considering different data subranges, from the entire range to smaller chunks. We choose to use Savitzky-Golay smoothing to identify crucial points in the data connected with changes in the character of the data, which can be used as borders between two clusters.

\begin{figure*}
\begin{minipage}{.75\textwidth}
\begin{algorithm}[H]
\caption{recursive\_clustering}\label{algclustering}
\begin{algorithmic}

\State \textbf{Inputs:} \var{data}, \var{hist}, \var{w\_length}, \var{poly}, \var{n\_iter}, \var{min\_n\_elem}
\State \textbf{Output:} the final division of the parameter range
\Function{recursive\_clustering}{\var{data}, \var{hist}, \var{w\_length}, \var{poly}, \var{n\_iter}, \var{min\_n\_elem}}
\State \var{n\_clust} := \Call{smoothing}{ \var{hist}, \var{w\_length}, \var{poly} }
\If {\var{n\_clust} == 1 \OR  \Call{len} { \var{hist} } $<$ \var{min\_n\_elem}}
	\State \Return 1
\Else

	\State \var{borders} := \Call{best\_clustering}{\var{n\_clust}, \var{data}, \var{n\_iter}}
	\State \var{hists} := \Call{subhists}{\var{hist}, \var{borders}}
	\State \var{datas} := \Call{subdatas}{\var{data}, \var{borders}}
    \State \var{division} := []
	\For {\var{w} in (0, \Call{len}{\var{borders}}) }
		\State \var{division}.\textsc{append}(\Call{recursive\_clustering}{\var{datas[w]}, \var{hists[w]}, \var{w\_length}, \var{poly}, \var{n\_iter}, \var{min\_n\_elem}})
	\EndFor
    \If{\Call{sum}{\var{division}} == \var{n\_clusters}}
        \State \Return \var{n\_clusters}
    \Else
        \State \Return \var{division}
    \EndIf
\EndIf
\EndFunction
\end{algorithmic}
\end{algorithm}
\end{minipage}
\end{figure*}

The Savitzky-Golay technique \cite{savitzky} is one of the most popular filters used for smoothing experimental data. It is based on the idea of local least-squares fitting data by polynomials. 
\begin{equation}
g_n=\sum_{i=-i_0}^{i_0} c_i f_{n+i},
\end{equation}
where $f$ denotes the original value, $g$ is the new value, and $c_i$ coefficients are obtained from fitting the assumed degree's polynomial and its derivatives.
It allows the minimization of the influence of random noise in the data under consideration.

\begin{figure*}
\begin{minipage}{.75\textwidth}
\begin{algorithm}[H]
\caption{best clustering (with the use of k-means)}
\begin{algorithmic}
\State \textbf{Inputs:} \var{n\_iter}, 
\State \textbf{Output:} the best (in the sum of square measure) distribution of centroids
\Function{best\_clustering}{\var{n\_clust}, \var{data},  \var{n\_iter}}
\State \var{best\_sum} := \var{MAX\_INT}
\State \var{best\_centroids} :=[]
\For {\var{i} in (0,\var{n\_iter})}
	\State \var{centroids} := \Call{k\_means}{ \var{n\_clust}, \var{data}}
	\State \var{sum} := \Call{sum\_of\_squares} {\var{centroids}, \var{data}}
	\If {\var{best\_sum}$>$\var{sum}}
		\State \var{best\_sum} := \var{sum}
		\State \var{best\_centroids} := \var{centroids}
	\EndIf
\EndFor
\State \Return \Call{centroids\_to\_borders}{\var{best\_centroids}}
\EndFunction
\end{algorithmic}\label{algkmeans}
\end{algorithm}
\end{minipage}
\end{figure*}

\begin{figure*}
\begin{minipage}{.75\textwidth}
\begin{algorithm}[H]
\caption{smoothing}
\begin{algorithmic}
\State \textbf{Inputs:} \var{hist}, \var{w\_length}, \var{poly}
\State \textbf{Output:} the number of clusters
\Function{smoothing}{\var{hist}, \var{w\_length}, \var{poly}}
\State \var{pos\_1} := \var{MAX\_INT}
\State \var{pos\_2} := \var{MAX\_INT}
\State \var{pos\_3} := \var{MAX\_INT}
\State \var{iter} := 0

\Repeat
 	 	\State \var{s\_hist} := \Call{SG\_filter} { \var{hist}, \var{w\_length}, \var{poly} }
		
		\If {\var{n\_hills} == 1 \AND \var{iter} $<$ \var{pos\_1}}
			\State \var{pos\_1} := \var{iter}
		\ElsIf {\var{n\_hills} == 2 \AND \var{iter} $<$ \var{pos\_2}}	
			\State \var{pos\_2} := \var{iter}
		\ElsIf {\var{n\_hills} == 3 \AND \var{iter} $<$ \var{pos\_3}}	
			\State \var{pos\_3} := \var{iter}	
		\EndIf	
		
		\var{iter} :+= 1
\Until  {\var{n\_hills} ( \var{s\_hist} ) $>$ 1}

\If{\var{pos\_1} == \var{MAX\_INT}}
	\State \var{n\_clust} := 1
\ElsIf{\var{pos\_2} \/ \var{pos\_3} $<$ \var{n\_buckets}}
 	\State \var{n\_clust} := 2	
\Else \State \var{n\_clust} := 3	
\EndIf
\State \Return \var{n\_clust}
\EndFunction
\end{algorithmic}\label{algsmoothing}
\end{algorithm}
\end{minipage}
\end{figure*}

We look at the data using a top-down approach and use the Savitzky-Golay filter at each recursion step to decide if a subrange should be divided into one, two, or three clusters. The recursive procedure of clustering is presented in the form of pseudocode Algorithms (\ref{algclustering} - \ref{algkmeans}). 
For simplicity of the presentation, we split the algorithm into three main parts: recursive scheme (Algorithm \ref{algclustering}),  
division of part of the data into subranges (Algorithm \ref{algkmeans}), and smoothing operation using S-G filter (Algorithm \ref{algsmoothing}). 

\section{Numerical experiments}

For all three mentioned earlier datasets, we performed the same procedure:
we use the proposed Savitzky-Golay recursive scheme combined with SOM and k-means clustering algorithms. We use the following set of parameters: the number of centers k $\in \{2,3,4,5\}$, learning rate $0.5$, change of learning rate $0.93$, change of center's potential $0.99$ for SOM (see \cite{tavan}), and initialization scheme k-means++ \cite{arthur}. The maximal number of iterations in one run set to 300, relative tolerance with regards to inertia (used as additional stop condition) $10^{-4}$. As the final result of clustering, we use the best one of 10 runs of the appropriate algorithm.

\subsection{Clustering of ground temperature measurements data}

\begin{figure}[!h!]
\centering
    \includegraphics[width=1.0\columnwidth]{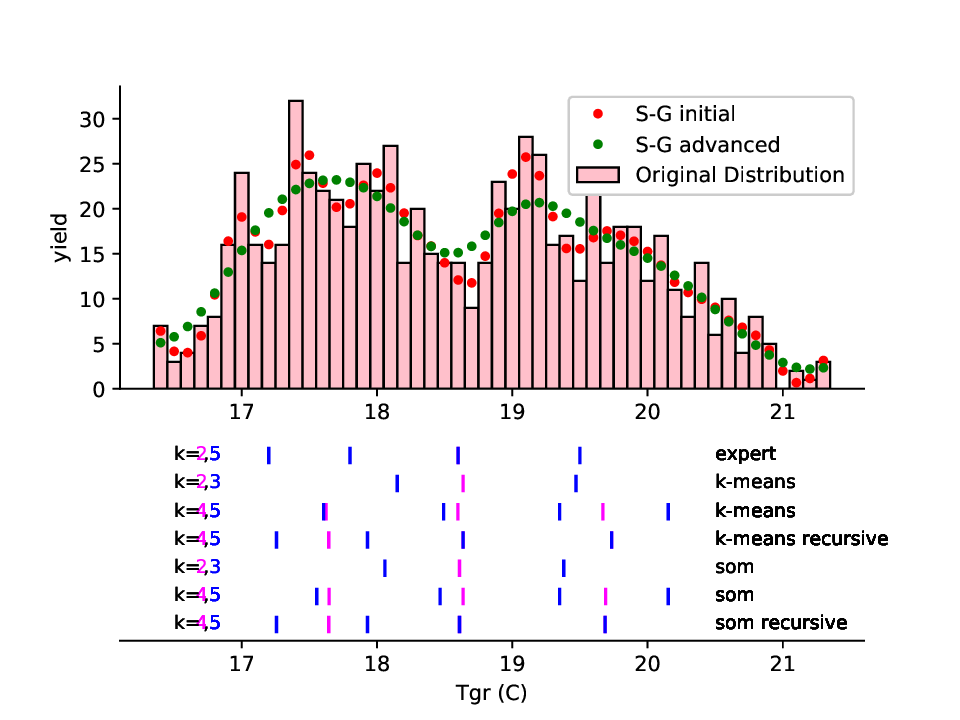}
    \caption{The results of different clustering methods applied to ground temperature data. The histogram data and the result of Savitzky-Golay (S-G) smoothing are presented at the top. Below, we marked cluster borders for different methods and different numbers of clusters.}\label{tgr_sg_hist}
\end{figure}

The first dataset studied is, as mentioned earlier, ground temperature.
The results for different clustering methods and numbers of clusters is presented in Figure \ref{tgr_sg_hist}. The histogram analysis suggests that the best division should be into two or five clusters. The histogram splits into two parts around a minimal point near 18.7. As the data on the left from this point is more sophisticated, according to experts, it should be divided into three parts, while the part on the right can be split into two subclusters. We can observe that the recursive scheme gives division into clusters very similar to the expected one, denoted as expert clustering, while all results obtained using out-of-the-box k-means and SOM approaches differ significantly.

Moreover, we measure the similarities between different clusterings of the same set of data using: Rand ($R$), adjusted Rand ($R'$), Jordan ($J$), Fowles-Mallows ($FW$), Arabie-Boorman ($AB$), and Hubert ($H$). The results are listed in Table \ref{miary}. All the selected measures are often used to compare similarities in two different clustering methods. Therefore, we show pairwise similarity between different clustering approaches. At first, we see that classical k-means and SOM methods give very similar results, as all measures show high relevance when the same number of clusters is chosen. For example, for four clusters obtained $R'$ between methods SOM[4] and k-means[4] equals 0.944. On the other hand, we can observe that the recursive approach SOM$_r$[3;2] differs significantly from classical SOM[5] as $R'$ values the value 0.491. In both these cases, the dataset is divided into five clusters, and we could expect the results to be similar. An additional important observation arises when comparing recursive k-means and recursive SOM approaches. Again, a high value of $R'$ suggests that it is unimportant which of these two we choose as a base for the presented recursive approach. Both produce very similar clusters.
\begin{table}[!h]
\centering
\caption{Measure of similarities between different clusterings. SOM$_r$ and k-means$_r$ denote the application of recursive methods. Numbers in the brackets describe the splitting into clusters.}
\footnotesize{

\begin{tabular}{|c|c|c|c|c|c|c|c|c|}
\hline
 &   & $R$ & $R'$ & $FM$ & $J$ & $AB$ & $H$ \\ \hline
SOM[5]&	SOM$_r$[3;2]&	0.832&	0.491&	0.597&	0.425&	0.168&	0.664\\ \hline
SOM[5]&	k-means[5]&	0.961&	0.884&	0.909&	0.833&	0.039&	0.922\\ \hline
SOM$_r$[3;2]&	k-means$_r$[3;2]&	0.953&	0.859&	0.888&	0.799&	0.047&	0.907\\ \hline
SOM$_r$[2;2]&	SOM[4]&	1&	1&	1&	1&	0&	1\\ \hline
SOM$_r$[2;2]&	k-means$_r$[2;2]&	0.978&	0.943&	0.957&	0.918&	0.022&	0.957\\ \hline
SOM[4]&	k-means[4]&	0.979&	0.944&	0.958&	0.920&	0.021&	0.958\\ \hline
SOM[3]&	k-means[3]&	0.919&	0.821&	0.883&	0.790&	0.081&	0.839\\ \hline
SOM[2]&	k-means[2]&	0.997&	0.994&	0.997&	0.994&	0.003&	0.994\\ \hline
\end{tabular}}
\label{miary}
\end{table}

\subsection{Clustering of water level measurements data}

\begin{figure}[!ht]
\centering
\includegraphics[width=1.0\columnwidth]{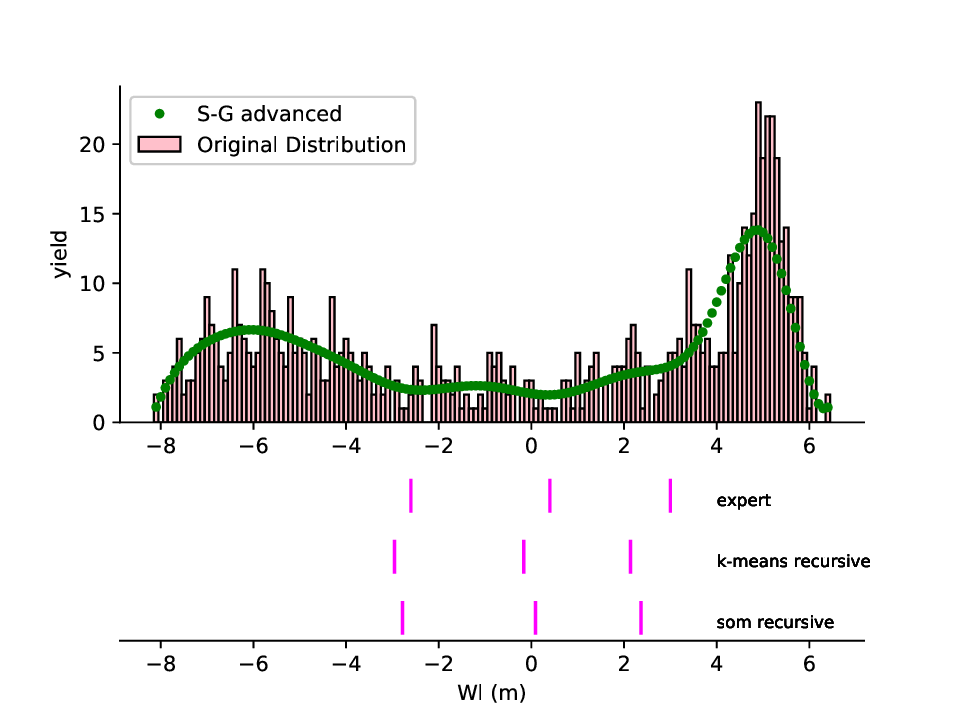}
\caption{Clustering of data for water level. The template of the picture is the same as in Fig.~\ref{tgr_sg_hist}. The borders between the clusters are shown in the lower part of the Figure.}
\label{lf_wl}
\end{figure}

The next test case is the water level measured simultaneously with the ground temperature discussed earlier. Fig.\ref{lf_wl} shows the data distribution, the results of clustering with SOM and k-means, and the expected division given by the expert. Once again, results obtained via k-means and SOM clustering are not satisfactory. Using the recursive Savitzky-Golay scheme, we can improve the clustering results. As in the previous case, the recursive scheme used with SOM and k-means gave similar clusters. 

\subsection{Clustering of Banknote authentication dataset}

In this subsection, we present the analysis of clustering of the data taken from the public repository. We decide to use one of the parameters describing the Banknote Authentication Dataset \cite{mlr} - the entropy of the image. The shape of the histogram differs from the previous ones since here we observe the long tail for low values and many narrow dips spread over the whole domain. Unlike earlier distributions, here we propose our subjective division into clusters, for similarity also described as ``expert''. We identify essential dips separating the intervals where the lack of smoothness can be connected to statistical effects. Intentionally, we do not divide the slopes of the highest peak. This procedure produces 7 clusters. Although several borders' locations are similar, when applying the SG-based technique, some additional clusters are recognized. This concerns especially the slopes, which are visibly divided into several clusters, and the different organization of the tail of the distribution.

\begin{figure}[!ht]
\centering
\includegraphics[width=1.0\columnwidth]{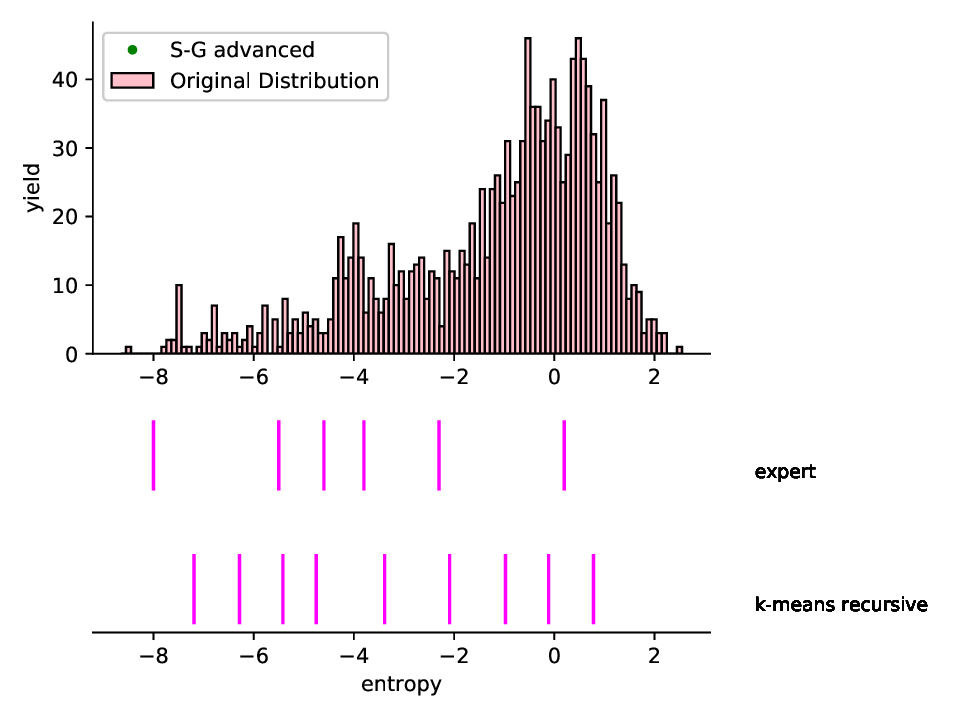}
\caption{Clustering of data for entropy of image dataset. The borders between the clusters are shown in the lower part of the Figure.}
\label{podzial_tbank}
\end{figure}

The results of each clustering approach are presented in Fig.\ref{podzial_tbank}.

\section{Conclusions}

The paper presents the recursive scheme for clustering that combines classical elements like k-means/SOM and Savitzky-Golay smoothing. The proposed approach has been applied to three different datasets. Where possible to compare, the clusters obtained via the proposed recursive scheme correspond better to the experts' estimations. For the Banknote dataset, some features of the method are better visible, especially the behavior on the distribution slopes. The crucial property of the process is also the automatic detection of the number of clusters. 
The analysis of clusterization of experimental data related to greenhouse gas exchange with many measurement errors.
The proposed recursive method gives distinctly better results than classical approaches.

Some further steps can be made to the proposed approach. These steps concern especially two issues: the detailed analysis of the determination of bin number selection and the parameters of SG filtering, as well as the better automation of the procedure.

\bibliographystyle{authordate1}

\begin{thebibliography}{9}
\bibitem{arthur}
                 Arthur, D. and Vassilvitskii, S.,
              Proceedings of the Eighteenth Annual ACM-SIAM Symposium on Discrete Algorithms,
                  1027-1035,
              Society for Industrial and Applied Mathematics,
                  K-means++: The advantages of careful seeding,
                   2007,
 

 \bibitem{brouwer}
                 Brouwer, R.K.,
                  of Intelligent Information Systems,
                  213-235,
                  Extending the Rand, adjusted Rand and Jaccard indices to fuzzy partitions,
                 32(3),
                   2009,
 

 \bibitem{everitt}
                 Everitt, B. and Landau, S. and Leese, M. and Stahl D.,
              Wiley,
                  Cluster Analysis, 5th Edition,
                   2011,
 

 \bibitem{fortuniak}
                 Fortuniak, K. and Pawlak, W. and Bednorz, L. and Grygoruk, M. and Siedlecki, M. and Zielinski, M.,
                Agricultural and Forest Meteorolog,
                  306-318,
                  Methane and carbon dioxide fluxes of a temperate mire in Central Europe,
                 232,
                   2017,
 

 \bibitem{fowlkesmallows}
                 Fowlkes, E.B. and Mallows, C.L.,
                  of the American Statistical Association,
                  553-56,
                  A method for comparing two hierarchical clusterings,
                 78(383),
                   1983,
 

 \bibitem{hubertarabie}
                 Hubert, L. and Arabie, P.,
                  of Classification,
                  193-218,
                  Comparing partitions,
                 2(1),
                   1985,
 

 \bibitem{kohonen}
                 Kohonen, T.,
                Proceedings of the IEEE,
                  1464-1480,
                  The self-organizing map,
                 78(9),
                   1990,
 

 \bibitem{lloyd}
                 Lloyd, S.P.,
                IEEE Transactions on Information Theory,
                  129-137,
                  Least squares quantization in PCM,
                 28(2),
                   1982,
 

 \bibitem{mlr}
  		Dua, Dheeru and Graff, Casey,
  		2017,
  		UCI Machine Learning Repository,
  		http://archive.ics.uci.edu/ml,
  University of California, Irvine, School of Information and Computer Sciences,
 

 \bibitem{macqueen}
                 MacQueen, J.,
                Proceedings of the Berkeley Symposium on Mathematical Statistics and Probability,
                  281-296,
                  Some methods for classification and analysis of multivariate observations,
                 1,
                   1967,
 

 \bibitem{moorepeleg}
                 Moore, A. and Pelleg, D.,
                Proceedings of the fifth ACM SIGKDD international conference on Knowledge discovery and data mining,
                  277-281,
                  Accelerating exact k -means algorithms with geometric reasoning,
                   1999,
 

 \bibitem{rand}
                 Rand, W.M.,
                  of the American Statistical Association,
                  846-850,
                  Objective criteria for the evaluation of clustering methods,
                 66(336),
                   1971,
 

 \bibitem{rodriguez}
                 Rodriguez, M.Z. and Comin, C.H. and Casanova, D. and Bruno, O.M. and Amancio, D.R.,
                PLoS ONE,
                  Clustering algorithms: A comparative approach,
                 14(1),
                   2019,
 

 \bibitem{savitzky}
                 Savitzky, A. and Golay, M. J. E.,
                Analytical Chemistry,
                  1627-1639,
                  Smoothing and differentiation of data by simplified least squares procedures,
                 36(8),
                   1964,
 

 \bibitem{tavan}
                 Tavan, P. and Grubmuller, H. and Kuhnel, H.,
                Biological cybernetics,
                  95-105,
                  Self-organization of associative memory and pattern classification: recurrent signal processing on topological feature maps,
                 64,
                   1990,
 

 \bibitem{upalneufeld}
                 Upal, M.A. and Neufeld E.,
                Proceedings of the First International Conference on Information, Statistics and Induction in Science,
                  342-353,
                  Comparison of unsupervised classifiers,
                   1996,
 

 \bibitem{warrens}
                 Warrens, M.J.,
                  of Classification,
                  177-183,
                  On the equivalence of Cohen's kappa and the Hubert-Arabie adjusted Rand index,
                 25(2),
                   2008,
 

 \bibitem{Rousseeuw_87}
    Rousseeuw, Peter J.,
    Silhouettes: A graphical aid to the interpretation and validation of cluster analysis,
      of Computational and Applied Mathematics,
    20,
    53 - 65,
    1987,
    0377-0427,



 \bibitem{Ketchen_96}
    Ketchen, David J. and Shook, Christopher L.,
    The application of cluster analysis in Strategic Management Research: An analysis and critique,
    Strategic Management  ,
    17,
    6,
    441-458,
    1996,



 \bibitem{Tibishirani_01}
    Tibshirani, Robert and Guenther, Walther and Hastie, Trevor,
      of the Royal Statistical Society Series B,
    Estimating the   of Clusters in a Data Set via the Gap Statistic,
    63,
    411-423,
    2001,


 \bibitem{Sugar_03}
    Sugar, Catherine A.  and James, Gareth M. ,
    Finding the   of Clusters in a Dataset,
      of the American Statistical Association,
    98,
    463,
    750-763,
     2003,
    Taylor and Francis,


 \bibitem{Lloyd_82}
   Lloyd, S., 
   IEEE Transactions on Information Theory, 
   Least squares quantization in PCM, 
   1982, 
   28, 
   2, 
   129-137,
 

\bibitem{MacQueen_67}
Berkeley, Calif.,
    MacQueen, J.,
    Proceedings of the Fifth Berkeley Symposium on Mathematical Statistics and Probability,   1: Statistics,
    281--297,
    University of California Press,
    Some methods for classification and analysis of multivariate observations,
    1967,


 \bibitem{Kohonen_82}
   Kohonen, Teuvo, 
   Biological Cybernetics, 
   Self-organized formation of topologically correct feature maps, 
   1982, 
   43, 
   2, 
   59-69,
 

 \bibitem{Jadhav_19}
    Jadhav, Amolkumar Narayan and N., Gomathi,
    EKEGWO: Enhanced Kernel-Based Exponential Grey Wolf Optimizer for Bi-Objective Data Clustering,
    International   of Uncertainty, Fuzziness and Knowledge-Based Systems,
    27,
    04,
    669-688,
    2019,


 \bibitem{Alguliyev_20}
    Alguliyev, Rasim M.  and Aliguliyev, Ramiz M.  and Sukhostat, Lyudmila V.,
    Weighted consensus clustering and its application to Big data,
    Expert Systems with Applications,
    150,
    113294,
    2020,
    0957-4174,


 \bibitem{Farshidvard_23}
    Farshidvard, A. and Hooshmand, F. and MirHassani, S.A.,
    A novel two-phase clustering-based under-sampling method for imbalanced classification problems,
    Expert Systems with Applications,
    213,
    119003,
    2023,
    0957-4174,


 \bibitem{Prakaisak_22}
    Prakaisak, Intouch and Wongchaisuwat, Papis,
    Article Hydrological Time Series Clustering: A Case Study of Telemetry Stations in Thailand,
    2022,
    Water (Switzerland),
    14,
    13,


 \bibitem{Carro-Calvo_21}
    Carro-Calvo, Leopoldo and Jaume-Santero, Fernando and GarcA­a-Herrera, Ricardo and Salcedo-Sanz, Sancho,
    k-Gaps: a novel technique for clustering incomplete climatological time series,
    2021,
    Theoretical and Applied Climatology,
    143,
    1-2,
    447--460,


 \bibitem{Atkinson_18}
    Atkinson, Anthony C. and Riani, Marco and Cerioli, Andrea,
    Cluster detection and clustering with random start forward searches,
    2018,
    Journal  of Applied Statistics,
    45,
    5,
    777--798,


 \bibitem{Hadi_22}
	  Hadi, Suha Mohammed and Alsaeedi, Ali Hakem and Dohan, Mohammed Iqbal and Nuiaa, Riyadh Rahef and Manickam, Selvakumar and Alfoudi, Ali Saeed D.,
	  Dynamic Evolving Cauchy Possibilistic Clustering Based on the Self-Similarity Principle (DECS) for Enhancing Intrusion Detection System,
	  2022,
	  International   of Intelligent Engineering and Systems,
	  15,
	  5,
	  252--260,


 \bibitem{Sui_22}
	  Sui, Jinping and Liu, Zhen and Liu, Li and Jung, Alexander and Li, Xiang,
	  Dynamic Sparse Subspace Clustering for Evolving High-Dimensional Data Streams,
	  2022,
	  IEEE Transactions on Cybernetics,
	  52,
	  6,
	  4173--4186,


 \bibitem{Colak_22}
	  Colak, Murat and Kaya, Ihsan and Karasan, Ali and Erdogan, Melike,
	  Two-phase multi-expert knowledge approach by using fuzzy clustering and rule-based system for technology evaluation of unmanned aerial vehicles,
	  2022,
	  Neural Computing and Applications,
	  34,
	  7,
	  5479--5495,


\end{thebibliography}

\end{document}